\documentclass[final]{cvpr}

\usepackage{times}
\usepackage{epsfig}
\usepackage{graphicx}
\usepackage{amsmath}
\usepackage{amssymb}

\usepackage{url}            
\usepackage{booktabs}       
\usepackage{amsfonts}       
\usepackage{nicefrac}       
\usepackage{microtype}      
\usepackage{xcolor}         
\usepackage{subcaption}
\usepackage{multirow}
\usepackage{colortbl}
\usepackage{threeparttable}
\usepackage{makecell}
\definecolor{cvprblue}{rgb}{0.21,0.49,0.74}
\usepackage[pagebackref,breaklinks,colorlinks,allcolors=cvprblue]{hyperref}

\def\confName{CVPR}
\def\confYear{2025}

\title{MDA: An Interpretable and Scalable Multi-Modal Fusion under Missing Modalities and Intrinsic Noise Conditions}

\author{Lin Fan\thanks{Equal Contribution}\\
Southwest Jiaotong University\\
Chengdu, China\\
\and
Yafei Ou$^*$\\
Tokyo Institute of Technology\\
Yokohama, Japan\\
{\tt\small ou.y.ac@m.titech.ac.jp}
\and
Cenyang Zheng\\
Southwest Jiaotong University\\
Chengdu, China\\
\and
Pengyu Dai\\
Tokyo Institute of Technology\\
Yokohama, Japan\\
\and
Tamotsu Kamishima\\
Hokkaido University\\
Sapporo, Japan\\
\and
Masayuki Ikebe\\
Hokkaido University\\
Sapporo, Japan\\
\and
Kenji Suzuki\\
Tokyo Institute of Technology\\
Yokohama, Japan\\
\and
Xun Gong\\
Southwest Jiaotong University\\
Chengdu, China\\
}

\begin{document}
\maketitle

\begin{abstract}
Multi-modal learning has shown exceptional performance in various tasks, especially in medical applications, where it integrates diverse medical information for comprehensive diagnostic evidence. However, there still are several challenges in multi-modal learning, 1. Heterogeneity between modalities, 2. uncertainty in missing modalities, 3. influence of intrinsic noise, and 4. interpretability for fusion result. This paper introduces the Modal-Domain Attention (MDA) model to address the above challenges. MDA constructs linear relationships between modalities through continuous attention, due to its ability to adaptively allocate dynamic attention to different modalities, MDA can reduce attention to low-correlation data, missing modalities, or modalities with inherent noise, thereby maintaining SOTA performance across various tasks on multiple public datasets.
Furthermore, our observations on the contribution of different modalities indicate that MDA aligns with established clinical diagnostic imaging gold standards and holds promise as a reference for pathologies where these standards are not yet clearly defined. The code and dataset will be available.

\end{abstract}

%

\section{Introduction}
Recently, multi-modal learning has demonstrated outstanding performance across various tasks \cite{huang2021makes}, particularly in medical applications, as it integrates information from different modalities, offering comprehensive diagnostic evidence for healthcare professionals~\cite{litjens2017survey}. It has been employed in tasks such as disease-assisted diagnosis~\cite{manh2022multi,jiao2024multi} and prognostic forecasting~\cite{ren2024cross}.

\begin{figure}[!t]
    \centering
    \includegraphics[width=\linewidth]{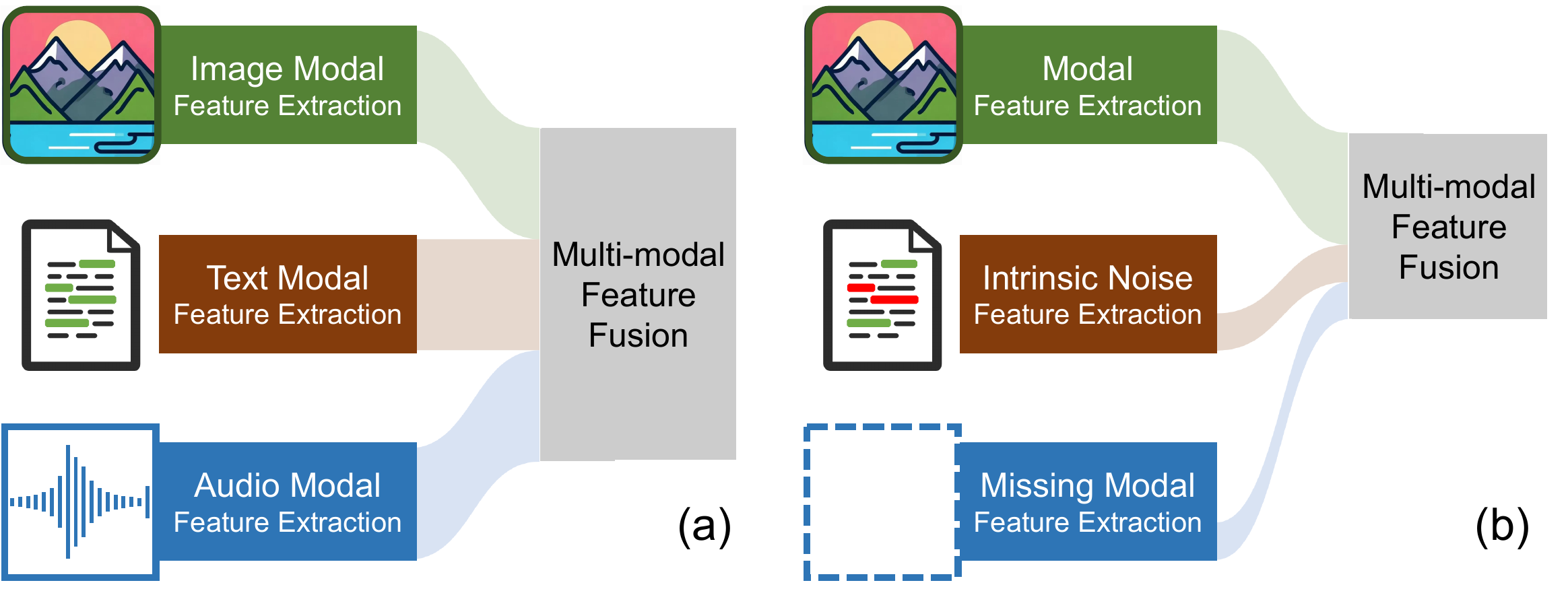}
    \caption{A unified multi-modal learning strategy involves learning with different multi-modal configurations. (a) Train and test with full modality. Different modalities receive equal attention (b) The model will reduce its modal-domain attention when learning with missing modalities or intrinsic noise.}
    \label{fig:1}
\end{figure}

Despite efforts in multi-modal medical research, fundamental and challenging issues remain due to the complexity of multi-modal clinical data and real-world application scenarios: 
\textbf{Chanllenge1: Multi-modal heterogeneity challenges fusion.} Modality heterogeneity includes information redundancy and complementary misalignment, which pose challenges for fusion. The greater the number of modalities, the more difficult the fusion process becomes. Various fusion methods have been proposed, initially based on multi-scale transformations ~\cite{jie2022tri,mo2021attribute} and sparse representations ~\cite{li2021joint,vishwakarma2018image}, and later using deep learning techniques \cite{ye2023f,zhang2023fdgnet,zhao2023cddfuse}. However, these methods fail to effectively exploit modality synergy, often relying on simple concatenation of latent features, allocating the same attention to each modality, which limits the potential of multi-modal integration. Recently, multi-modal attention mechanisms have been introduced \cite{sun2023attention, ma2023multi}, but they primarily focus on mutual attention between two modalities. A key limitation of the above methods is their inability to dynamically allocate attention based on the importance of each modality to the final diagnosis, especially in the case of multiple heterogeneous multi-modal data, as shown in Fig.~\ref{fig:1} (a). 
\begin{figure}[!t]
    \centering
    \includegraphics[width=\linewidth]{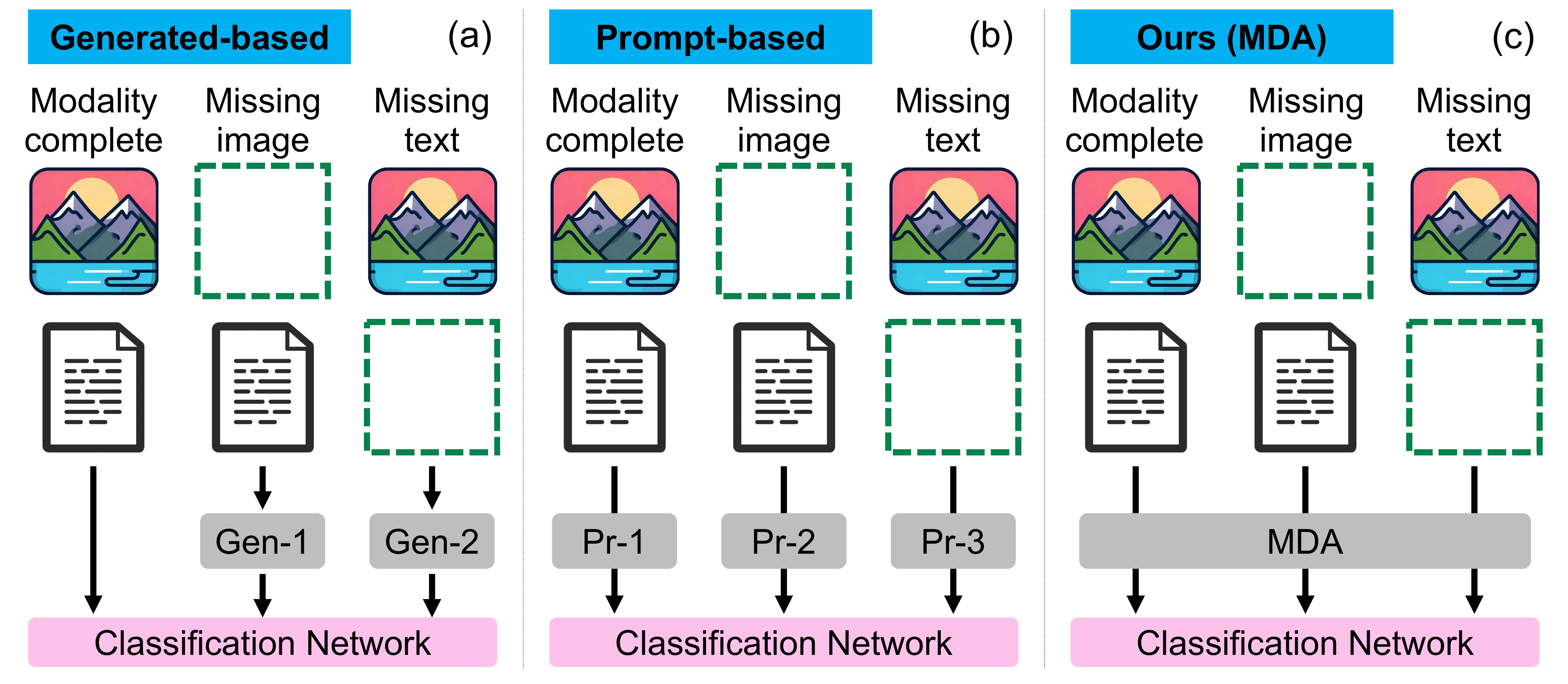}
    \caption{Existing methods vs. MDA for missing modality. Generation-based methods ~\cite{cui2022survival} add specific models (Gen-1, Gen-2) for missing modalities, similar to prompt-based methods ~\cite{lee2023multimodal} that add different prompts (Pr-1,Pr-2, Pr-3), complexifying network parameters intensify training difficulty. 
    This work builds linear attention relationships between modalities, adaptively adjusting their weights in real-time, and more efficiently handling various scenarios involving missing modalities.}
    \label{fig:scale}
\end{figure}
\begin{figure*}[!t]
    \centering
    \includegraphics[width=\textwidth]{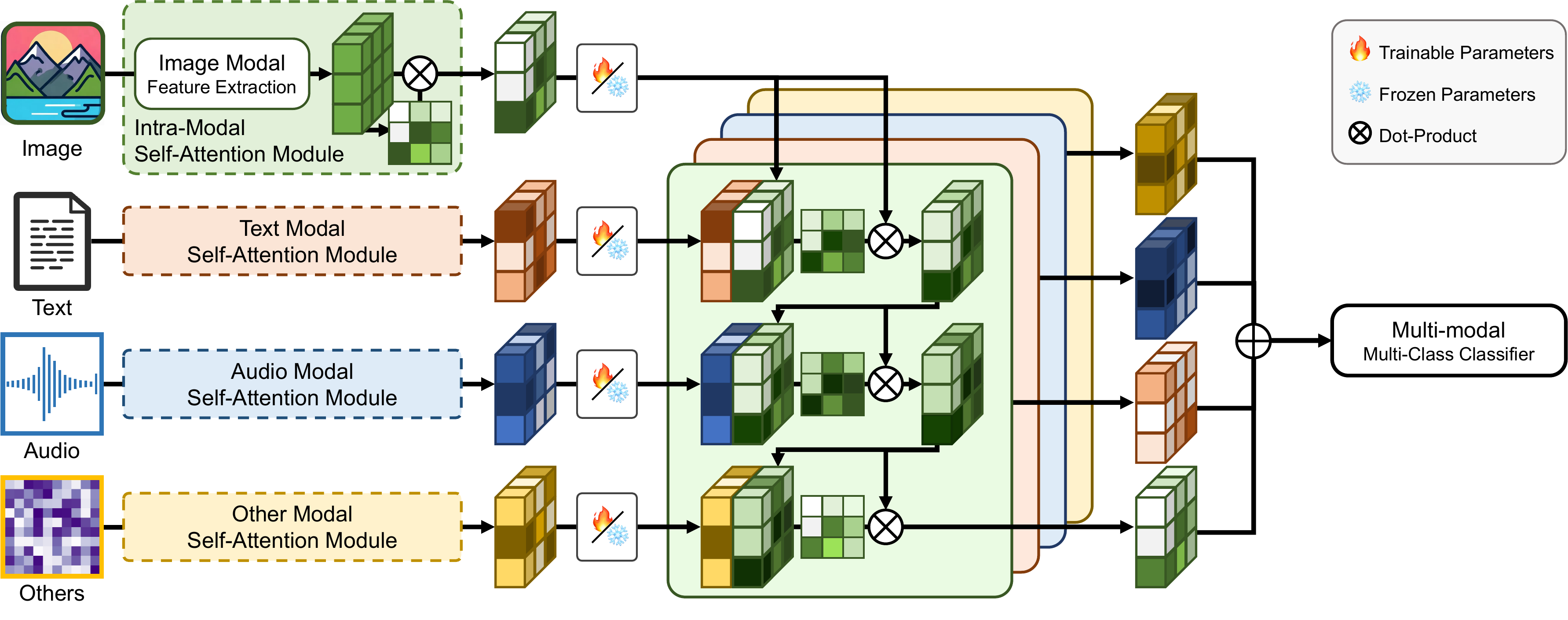}
    \caption{Overview of the proposed network framework. The uni-modal training involves building pre-trained models for multi-disease classification across different modalities, with a dedicated classifier assigned to each disease. The work introduces the weight calculation for each scenario under the proposed modal-domain attention module.}
    \label{fig:2}
\end{figure*}

\textbf{Chanllenge2: The issue of modality missing in real-world scenarios.} 
Missing modality is an inevitable issue in medical multi-modal learning. Several generate-based methods \cite{yang2022robust,li2023incomplete,ren2024novel} are dedicated to designing distinct generative models for each missing modality to achieve full modality effects, as shown in Fig.~\ref{fig:scale} (a). Recently, a series of approaches including MAPs \cite{lee2023multimodal,dai2024muap,jang2024towards} fine-tuning the Visual-Language model by constructing specialized prompts for each missing modality to address the issue, as shown in Fig.~\ref{fig:scale} (b). While these prompt-based methods offer a more flexible and scalable solution than generate-based methods. However, the inherent complexity of designing and maintaining a large number of generative networks or prompts increases with the number of modalities~\cite{hu2021model}. Moreover, the assumption that a dedicated generative network can effectively handle each missing modality or prompt may not always hold, especially in scenarios where the relationships between modalities are complex and non-linear.

\textbf{Chanllenge3: Learning with intrinsic noise.} 
As the field of multi-modal research gains momentum, the intra-modal cross-verification of information presents innovative perspectives for noise learning~\cite{liu2024minjot}, rather than focusing on label noise~\cite{karimi2020deep,wang2024simple,penso2024confidence}. 
The exploration of intrinsic noise within text data remains limited in multi-modal research. Intrinsic noise, that is, discrepancies between textual information and clinical realities, can introduce contradictions during modality fusion. 
This phenomenon is particularly pronounced in the medical domain, where diagnostic reports are often laden with subjective annotations made by clinical experts, as highlighted by Fusaroli~\cite{fusaroli2021quality}. Exploiting the complementary differences between modalities to mitigate the impact of intrinsic noise may serve as an effective solution in multi-modal fusion. 

\textbf{Chanllenge4: The comprehensive interpretability of multi-modal fusion.} 
Currently, techniques such as heat maps are widely used to improve the interpretability of models, allowing clinicians to gain direct insights into the regions underlying diagnostic decisions~\cite{manh2022multi,wu2023expert}. However, these methods are limited to providing explanations within the confines of individual modalities and do not extend to cross-modal scenarios. The interactive interpretability between multi-modal lacks a unified paradigm, which is precisely what is needed in clinical practice. For instance, comprehensive diagnosis of gastrointestinal tumors in clinical settings often utilizes multi-modal imaging techniques such as EUS, WLE, CT, and MRI, with each modality being sensitive to different disease types ~\cite{jacobson2023acg}. Some diseases have established diagnostic gold standards, such as WLE for lipomas. However, many other conditions still lack definitive diagnostic criteria. The medical community is increasingly focused on establishing diagnostic gold standards, aiming to standardize procedures and reduce the number of imaging modalities required ~\cite{richardson1995well, chandler2019cochrane, jacobson2023acg}. Therefore, exploring how to construct a paradigm for interactive interpretability between multi-modalities has become an urgent task in the field of medical image analysis.

We argue that the essence of the challenges above is uniform, focusing on how to efficiently and flexibly process and integrate multi-modal data, each carrying distinct information. 
We aim to address the above challenges by adjusting the weight of each modality in the final decision-making process.
Guided by this, we introduce the Modal-Domain Attention (MDA), which employs an innovative continuous attention algorithm, establishing a linear attention relationship between multi-modal features, flexibly allocating attention to different modalities: it reduces the attention to modalities with lower relevance to the final decision, to missing modalities, and to those containing intrinsic noise, as illustrated in Fig.~\ref{fig:1} (b), and can more efficiently handle various scenarios involving missing modalities, as illustrated in Fig.~\ref{fig:scale} (c). 
Our contributions are as follows:

\begin{itemize}
\item MDA can adaptively adjust the weight of each modality in the final decision-making process and has achieved SOTA performance across multiple multi-modal datasets, demonstrating that this strategy can effectively enhance performance in multi-modal tasks.
\item MDA can adjust the attention for each modality in decision-making, reducing the emphasis on low-relevance information and minimizing the impact of missing modalities and intrinsic noise on outcomes.
\item Based on MDA, this study comprehensively analyzes the sources of multi-modal efficacy from both macroscopic (across different diseases) and microscopic perspectives (within individual cases), while pioneering in exploring the interpretability of multi-modal learning as the gold standard for clinical diagnostics, thereby promoting the precision and standardization of medical diagnostics for standards-lacking diseases.
\end{itemize}

\section{Methodology}
This study introduces MDA, an adaptable multi-modal fusion strategy. MDA fortifies against the missing of any modality and reduces intrinsic noise influence by adjusting the modal contributions to decision-making. These contributions index the decision's reliance on each modality. The overview of the method is shown in Fig.~\ref{fig:2}. Given ${M}$ image modalities ${I}$, ${P}$ text modalities ${T}$, ${Q}$ audio modalities ${A}$. The total number of modalities is ${N (N = M + P + Q)}$. The multi-modal dataset ${\mathcal{X}}$ can be represented as $\mathcal{X}=\{\{\mathbf{I}_j\}_{j=1}^M,\{\mathbf{T}_k\}_{k=1}^P,\{\mathbf{A}_l\}_{l=1}^Q\}$, with labels ${Y}$. Then Individual classification tasks ${\mathcal{T}_j=\{\mathbf{I}_j, Y\}}$ are defined for each modal, with analogous definitions for others. Finally, features are processed by MDA to determine adaptive weights. The final classification is achieved by fusing modal features based on these weights and inputting them into a classifier. The multi-modal classification task is defined as ${\mathcal{T}=\{\mathcal{X}, Y\}}$, where ${Y}$ is a one-hot vector for multi-class classification. Subsequent sections detail the processing of each modality.

\subsection{Image modalities}
Pre-trained convolutional neural networks (CNNs) are utilized as feature extractors for image modalities.
Take image modality $\mathbf{I}_j$ as an example, the vectorized latent features ${\mathbf{f}_j}$ obtained from the feature extractors ${\mathbf{G}_{j}}$ are denoted as:
\begin{equation}
{\mathbf{f}_j}=\mathbf{G}_j\left(\mathbf{I}_j\right),
\end{equation}
 
Multi-modal data fusion requires attention to inter-modal learning mechanisms and individual modality learning efficiency, both crucial for effective data fusion. We introduce self-attention to enable deeper feature exploration, enhancing cross-modal fusion and complementing inter-modal learning.
For image features, the calculation of self-attention $\mathbf{SA}$ scores is as Eq.~\ref{eq:phij}, 
where ${Q_j}$, ${K_j}$, and ${V_j}$ are linear transformations of the latent feature $\mathbf{f}_j$, ${\mathbf{\Phi}_j}$ represents the self-attention features of the ${j}$ modal.
\begin{equation}
\mathbf{\Phi}_j=\mathbf{SA}(\mathbf{f}_j)=\operatorname{softmax}\left(\frac{Q_j \cdot (K_j)^T}{\sqrt{d}}\right) \cdot V_j
\label{eq:phij}
\end{equation}

\subsection{Text modalities}
We employ BERT~\cite{devlin2018bert} to encode the input text. BERT learns contextual representations of words or subwords in a text by using a self-attention mechanism. This is consistent with the self-attention module in image feature extraction. To adapt BERT to our specific task, we unfroze the last four layers of BERT for training. The self-attention feature extraction $\mathbf{\Phi}_k$ representation for textual reports $\mathbf{T}_k$ is as follows:
\begin{equation}
\mathbf{\Phi}_k=\mathbf{BERT}\left(\mathbf{T}_k\right)
\end{equation}

\subsection{Audio modalities}
We utilize Mel-frequency cepstral coefficients (MFCCs) \cite{tzanetakis2002musical} as the representation for the audio modality. MFCCs have gained extensive application in the field of audio and speech processing due to their ability to mimic human auditory perception, facilitate dimensionality reduction of features, exhibit robustness to variations, and demonstrate broad applicability. We also applied self-attention to the extracted audio features, taking audio modality ${\mathbf{A}_l}$ as an example, and its feature representation $\mathbf{\Phi}_l$ obtained after processing through the self-attention mechanism is denoted as:
\begin{equation}
\label{eq:mfcc}
\mathbf{\Phi}_l=\mathbf{SA}(\mathbf{f}_l)=\mathbf{SA}(\mathbf{MFCCs}\left(\mathbf{A}_l\right))
\end{equation}

\subsection{Continuous attention mechanism}
\label{section2.4}
Existing multi-modal attention mechanisms are typically designed for bi-modal fusion, focusing on point-to-point interactions between modalities, overlooking the collective influence of all other modalities. We assert that in real-world multi-modal data processing, the information or semantics of one modality are inherently intertwined with those of all other modalities, necessitating a more holistic and scalable fusion technique. In contrast, MDA aims to compute the attention for each modality while taking into account the influence of all other modalities, thereby achieving a comprehensive understanding of multi-modal information. This approach also enhances the scalability of the expansion of modalities in any given scenario. 
It achieves this by employing a continuous attention mechanism that constructs linear attention relationships among multi-modalities, allowing it to dynamically adjust the contribution of each modality to the decision-making process. This reduces the impact of modalities containing invalid or erroneous information, thereby enhancing the overall accuracy. The specific computational approach is illustrated as Eq.~\ref{eq:thetaen}, 
where \(\mathbf{\Theta}_i^{N-1}\) represents the attention map computed between modality \(i\) and the other $N-1$ modalities, which is also known as the MDA map \(\mathbf{\Theta}_i\). \( R \) denotes the ReLU activation.
\begin{equation}
\mathbf{\Theta}^e_i=
\begin{cases}
\operatorname{R}\left(\frac{ Q_i\cdot (K_{e})^T}{\sqrt{d}}\right)\cdot V_{i}^e, e=1\\
\operatorname{R}\left(\frac{ \mathbf{\Theta}^{e-1}_{i}\cdot (K_{e})^T}{\sqrt{d}}\right)\cdot V_{i}^e, 1\textless e \leq N-1
\end{cases}
\label{eq:thetaen}
\end{equation} 

MDA allows for infinite expansion without being limited to handling only two or three modalities, and the network complexity remains consistently at a low level. This scalability is crucial for applications that require continuous integration of new modalities over time, particularly in medical imaging, where the collection of more modal data is necessary to provide accurate and trustworthy judgments.
\subsection{Objective function}
For the classification training of single-modality model $i$, given the features ${\mathbf{\Phi}_i}$ obtained after the self-attention module, we employ a multi-layer perceptron classifier ${\mathcal{C}_s}$ for disease prediction. The model training is guided by the cross-entropy loss, defined as Eq.~\ref{eq:elli}. For the multi-modal classification training, the classification loss is defined as Eq.~\ref{eq:ell}, where ${\mathcal{C}_f}$ stands for the multi-modal classifier. 
\begin{equation}
\ell_i=\text { CrossEntropy }\left(\mathcal{C}_s\left(\mathbf{\Phi}_i\right), Y\right)
\label{eq:elli}
\end{equation}
\begin{equation}
\ell=\text { CrossEntropy }\left(\mathcal{C}_f\left(\mathbf\{\Theta_i^{N-1}\}_{i=1}^N\right), Y\right)
\label{eq:ell}
\end{equation}

\subsection{Training setting}
We have directly developed a model that operates under conditions of missing modalities and intrinsic noise. We processed the individual modality features obtained to simulate these scenarios. Specifically, we randomly select a certain proportion of samples from each batch of data and apply the following treatments: 1. randomly set the features of one modality to 1 to simulate the arbitrary missing modality, and 2. replace an attribute within the text modality with another random description to simulate intrinsic noise. 3. The rates of missing modality or inherent noise in the training and test sets is identical.

We employ the Adam optimizer with a weight decay rate of 1e-4 to optimize the model parameters. All experiments were conducted using the PyTorch framework with the Geforce RTX 4090 GPU.
\section{Datasets}
We conducted experiments using four different multi-modal datasets, encompassing diverse dimensions such as medical data, natural images, movies, and audio, to assess the generality of the proposed model comprehensively.
\begin{table}[!t]
\centering
\centering
\caption{Detail of the gastrointestinal disease dataset.}
\resizebox{\linewidth}{!}{

\begin{tabular}{>{\centering\arraybackslash}p{1.5cm}>{\centering\arraybackslash}p{1.5cm}>{\centering\arraybackslash}p{1.5cm}>{\centering\arraybackslash}p{1.5cm}>{\centering\arraybackslash}p{1.5cm}}
\toprule
& \# Cases & \# EUS & \# WLE & \# Reports \\
\midrule
GIST &149  &1539  & 460 &149 \\ 
GLM &117  &1123  &340  &117 \\
NET &107  &1187  &404  &107 \\
EP &18  &182  &65  &18 \\
Lipoma &16  &180  & 61 &16 \\
GS &5  &77  &18  &5 \\
PC &6  &65  &14  &6 \\
IFP &3  &24 &8  &3 \\ \midrule
Overall &421  &4213  &1370  &421 \\
\bottomrule
\end{tabular}}
\label{data}
\end{table}

\textbf{Gastrointestinal disease dataset}
We compiled a multi-center gastrointestinal disease dataset comprising three modalities: Endoscopic Ultrasound (EUS), White Light Endoscopy (WLE), and imaging reports. The dataset was gathered from two separate centers (***hospital and ***hospital) and captured using Olympus UM-2R and UM-3R equipment, totaling 421 cases spanning eight distinct categories (Gastrointestinal Stromal Tumor (GIST), Gastric Leiomyoma (GLM), Neuroendocrine Tumor (NET), Ectopic Pancrea (EP), Lipoma, Gastrointestinal Schwannomas (GS), Pneumatosis Cystoid (PC), and Inflammatory Fibroid Polyp (IFP)). Among these patients, the gender distribution is 53\% male to 47\% female, with an age range spanning from 27 to 78 years. The imaging reports (in Chinese) offer detailed descriptions of lesion characteristics observed within the two imaging modalities. The detailed data quantities for each category are shown in Table~\ref{data}. We divided the dataset into training and test sets based on the variability of the centers, encompassing 327 and 94 cases, respectively.

\textbf{Derm7pt dataset}~\cite{kawahara2018seven} is a multi-modal dataset that contains clinical, dermoscopic, and meta-data models for melanoma skin lesions in dermatology. We predict disease diagnosis (DIAG) and all 7 dermoscopic concepts, including blue-whitish veil (BWV), dots and globules (DaG), pigmentation (PIG), pigment network (PN), regression structures (RS), streaks (STR), and vascular structures (VS) in considered datasets. The dataset contains a total of 1011 cases. 

\textbf{Multi-modal IMDb (MM-IMDb)}~\cite{arevalo2017gated} contains two modalities: image and text. We conduct experiments on this dataset to predict a movie genre using
image and text modality, which is a multi-label classification task as multiple genres could be assigned to a single movie. The dataset includes 25, 956 movies and 23 classes. The division rule for this dataset adheres to the contrasting method SMIL~\cite{ma2021smil}.

\textbf{Audiovision-MNIST}~\cite{vielzeuf2018centralnet} consists of an independent image and audio modalities. The images, which are digits from 0 to 9, are collected from the MNIST dataset \cite{lecun1998gradient} with a size of 28 × 28, and the audio modality is collected from Free Spoken Digits Dataset containing raw 1500 audios. The dataset contains 1500 samples for both image and audio modalities. The division rule for this dataset adheres to the contrasting method SMIL~\cite{ma2021smil}.

\section{Experiments and discussion}
In our experiments, each uni-modal task is trained independently, with its parameters frozen during multi-modal fusion training. This approach ensures that the self-attention module of each uni-modal model does not interfere with the fusion module’s learning process, allowing any accuracy improvements to be attributed solely to the different multi-modal fusion strategies.
\begin{table}[!t]
\centering
\caption{Ablation experiments of multi-modal fusion assessing diagnostic accuracy (\%) under conditions of missing modalities and intrinsic noise.}
\resizebox{\linewidth}{!}{
\begin{threeparttable}
\begin{tabular}{>{\arraybackslash}p{2.25cm}|>{\centering\arraybackslash}p{2.25cm}|>{\centering\arraybackslash}p{1cm}|>{\centering\arraybackslash}p{1cm}|>{\centering\arraybackslash}p{1cm}}
\toprule
& Fusion & \multicolumn{3}{c}{Intrinsic Noise} \\ \cline{3-5}
Modal & Method & w/o & 5\% & 20\% \\ \midrule
Uni-EUS & - & 85.3 & - & - \\
Uni-WLE & - & 81.7 & - & - \\
Uni-Report & - & 87.4 & 85.7 & 80.2 \\ \midrule
EUS+WLE    & MDA (Ours) & 91.6 & - & - \\
EUS+Report & MDA (Ours) & 95.2 & - & - \\
WLE+Report & MDA (Ours) & 94.7 & - & - \\ \midrule
\multirow{4}{*}{\makecell[l]{Complete\\Modalities}}
& Addition & 87.6 & 87.0 & 85.5 \\
& Maximum & 88.3 & 88.2 & 86.7 \\
& Concatenate & 91.2 & 89.3 & 87.8 \\
& MDA (Ours) & \textbf{98.9} & \textbf{98.6} & \textbf{98.5} \\ \midrule
\multirow{4}{*}{\makecell[l]{70\%\\Modalities}} 
& Addition & 87.3 & 85.6 & 84.1 \\
& Maximum & 87.4 & 86.3 & 85.6 \\
& Concatenate & 89.7 & 88.4 & 87.9 \\
& MDA (Ours) & \textbf{98.6} & \textbf{98.4} & \textbf{97.3} \\ \midrule
\multirow{4}{*}{\makecell[l]{50\%\\Modalities}} 
& Addition & 85.7 & 82.8 & 81.9 \\
& Maximum & 84.2 & 81.1 & 80.2 \\
& Concatenate & 89.1 & 86.3 & 85.2 \\
& MDA (Ours) & \textbf{97.4} & \textbf{97.1} & \textbf{96.5} \\ \midrule
\multirow{4}{*}{\makecell[l]{20\%\\Modalities}} 
& Addition & 82.3 & 80.2 & 79.6 \\
& Maximum & 81.9 & 80.4 & 79.3 \\
& Concatenate & 84.7 & 82.1 & 81.0 \\
& MDA (Ours) & \textbf{95.6} & \textbf{95.7} & \textbf{94.6} \\ 
\bottomrule
\end{tabular}
\end{threeparttable}}
\label{table:1}
\end{table}

\begin{table*}[!t]
\centering
\caption{Comparison of classification results to the state-of-the-art multi-modal fusion methods in terms of accuracy (\%) by using Derm7pt~\cite{kawahara2018seven} dataset. The best values are highlighted in bold, and the second-best ones are highlighted underlined.}
\resizebox{\linewidth}{!}{
\begin{threeparttable}
\begin{tabular}{>{\arraybackslash}p{2.75cm}>{\centering\arraybackslash}p{1.25cm}>{\centering\arraybackslash}p{1.25cm}>{\centering\arraybackslash}p{1.25cm}>{\centering\arraybackslash}p{1.25cm}>{\centering\arraybackslash}p{1.25cm}>{\centering\arraybackslash}p{1.25cm}>{\centering\arraybackslash}p{1.25cm}>{\centering\arraybackslash}p{1.25cm}>{\centering\arraybackslash}p{1.25cm}}
\toprule
Method & BWV & DaG & PIG & PN & RS & STR & VS & DIAG & avg. \\ \midrule
Baseline~\cite{kawahara2018seven} & 87.1 & 60.0 & 66.1 & 70.9 & 77.2 & 74.2 & 79.7 & 74.2 & 73.7 \\
HcCNN~\cite{bi2020multi} & 87.1 & 65.6 & 68.6 & 70.6 & 80.8 & 71.6 & \underline{84.8} & 69.9 & 74.9 \\
FM4-SS~\cite{tang2022fusionm4net} & \underline{88.1} & 66.1 & 70.1 & 71.1 & 81.5 & \textbf{78.0} & 81.8 & 78.5 & 77.0 \\
RemixFormer~\cite{xu2022remixformer} & - & - & - & - & - & - & - & 81.3 & - \\
CAFNet~\cite{he2023co} & 87.8 & 61.5 & \textbf{73.4} & 70.1 & \underline{81.8} & \underline{77.0} & 84.3 & 78.2 & 76.8 \\
TFormer~\cite{zhang2023tformer} & 86.7 & \underline{66.9} & 70.3 & \underline{74.3} & \textbf{82.1} & 76.7 & 83.0 & 79.5 & \underline{77.4} \\
MICA~\cite{bie2024mica} & 84.4 & \textbf{79.1} & 68.8 & \textbf{74.4} & 72.8 & 71.3 & 81.3 & \underline{83.9} & 77.0\\
RGMC50~\cite{li2024self} & - & - & - & - & - & - & - & 78.9 & - \\
MDA (Ours) & \textbf{91.6} & \underline{66.9} & \underline{70.4} & 72.4 & 79.8 & 75.8 & \textbf{85.2} & \textbf{85.7} & \textbf{78.5} \\
\bottomrule
\end{tabular}
\end{threeparttable}}
\label{table:jbhi}
\end{table*}

\subsection{The efficacy of MDA in confronting the three key challenges of multi-modal fusion}
\label{sec:efficacy}
We present the results of our proposed method in addressing the challenges of multi-modal fusion, random missing modality, and learning with intrinsic noise in our gastrointestinal disease dataset. Table~\ref{table:1} shows the efficacy of our method for the three challenges via extensive analysis of ablation test results across three fusion scenarios.

\textbf{Scalability for multi-modality integration challenge.} The experimental results underscore the importance of multi-modal integration in enhancing diagnostic accuracy. Initially, when considering uni-modal approaches, the performance is limited, with individual modalities such as Uni-EUS and Uni-Report achieving accuracies of 85.3\% and 87.4\%, respectively. This highlights the inherent limitations of relying on a single data stream for diagnosis. As we progress to multi-modal fusion, the accuracy significantly improves, indicating the complementary nature of different modalities. For instance, the simple Concatenate fusion method already shows an enhancement, with an accuracy of 91.2\% when all modalities are present, surpassing the accuracy of any single modality.

As the number of integrated modalities increases, the MDA method exhibits a consistent and substantial improvement in accuracy. With the addition of each modality, the MDA's accuracy climbs from 95.2\% with two modalities to 98.9\% with three modalities, demonstrating its ability to effectively integrate and leverage information from multiple sources without increasing the computational load associated with branching pathways. This sustained attention mechanism in MDA is pivotal for achieving superior sustainable scalability in multi-modal data processing. This robustness in scalability, combined with its efficient approach to attention, allows MDA to excel in precision and scalability, making it a more effective method for integrating and processing multi-modal data in real-world scenarios.

\textbf{Missing modality challenge.} 
The sensitivity of multi-modal fusion methods to missing modalities is a critical factor in their practical application. Traditional fusion techniques such as concatenation or Addition often struggle to maintain performance when faced with incomplete data. In contrast, MDA exhibits remarkable stability in the face of missing modalities. The impact of missing data on MDA’s performance is substantially mitigated by its advanced attentional mechanism. When 70\% of the modalities are present, MDA still achieves an impressive accuracy of 98.6\%, which is a mere 0.3\% decrease from its performance with all modalities. This minimal change in accuracy, even with a large proportion of modalities missing. Further examining the resilience of MDA, we observe that as the percentage of missing modalities increases, the performance of other fusion methods tends to degrade more significantly. For example, if 80\% of one modality is randomly missing, the concatenate fusion accuracy might drop to 84.7\%, while MDA could still maintain an accuracy close to 95.6\%. This stark contrast in performance underscores the superior stability of MDA when dealing with substantial data gaps.

The stability of MDA in the face of missing modalities is not just a matter of maintaining high accuracy; it also reflects the method’s ability to effectively weigh and integrate the available information. The attention mechanism in MDA is designed to dynamically adjust to the presence or absence of modalities, ensuring that the model can leverage the most relevant and informative parts of the data. This adaptability is crucial in real-world diagnostic settings where the completeness of data cannot be guaranteed, and MDA’s performance under these conditions makes it a standout choice for multi-modal data integration.

\textbf{Learning with intrinsic noise challenge.} 
Table~\ref{table:1} highlights the resilience of MDA to intrinsic noise. While the accuracy of the Uni-Report modality decreases from 87.4\% without intrinsic noise to 80.2\% with 20\% noise, MDA maintains an impressive accuracy of 98.5\% with the same level of noise. This is a stark contrast to the concatenate method, which sees a decrease from 91.2\% to 87.8\% with 20\% noise. The MDA method’s ability to establish correspondences between modalities and effectively diminish the contribution of noisy modalities to the decision-making process is evident, as it restores diagnostic accuracy to a level comparable to that achieved with noise-free data (98.5\% vs. 98.9\%).

The analysis of the interplay between missing modalities and inherent noise has illuminated the resilience of different fusion techniques in the face of intricate modal forms. The results show that as the proportion of missing modalities increases, all methods experience a decline, but MDA consistently outperforms the others, even in noisy conditions. For example, with 50\% missing modalities and 5\% noise, MDA achieves 97.1\%, while traditional methods drop to 82.8\%, 81.1\%, and 86.3\%. Even with 20\% noise, MDA’s accuracy only slightly decreases to 96.5\%, while others see sharper declines. MDA’s superior performance is due to its advanced continuous attention algorithms, which allow it to handle incomplete and noisy data more effectively than traditional methods. As noise levels rise, the performance gap widens, particularly with 50\% and 20\ modality loss. The sharp decline in the efficacy of traditional methods highlights their vulnerability to noise when faced with substantial gaps in the data, whereas MDA demonstrates a more stable and resilient performance, underscoring its superior ability to maintain diagnostic accuracy even in the presence of substantial noise and missing data. This pronounced discrepancy underscores the robustness of MDA in handling complex data challenges, reaffirming its advantage over traditional approaches in the realm of multi-modal data integration.

\subsection{Comprehensive comparison in different database}
We conducted experiments on publicly available datasets to assess the efficacy of our proposed method in both complete and missing modal fusion scenarios. We benchmarked our approach against SOTA methods tailored for handling missing modalities. Across a range of tasks and various databases, MDA demonstrated SOTA performance.

\textbf{Comparison with related multi-modal fusion works}
The comparison results on the Derm7pt dataset are shown in Table~\ref{table:jbhi}. The table reports the quantitative results of the classification accuracy for each concept, highlighting the performance of MDA on various tasks. MDA outperforms other methods in most metrics, particularly in disease diagnosis, where it achieves an accuracy of 85.7\%, significantly surpassing the second-best method by 1.8\%. This exceptional performance in disease diagnosis demonstrates the robustness and effectiveness of MDA in integrating and analyzing multi-modal data, which is crucial for accurate diagnosis in medical studies.

\textbf{Comparison with related missing modality works}
\label{sec:comparison}
Due to the extensive research on missing modalities, we specifically conducted an extended investigation into the performance of MDA in the presence of missing modalities. Experiments were performed on the MM-IMDb and avMNIST datasets. 

\begin{table}[!t]
\centering
\caption{Model performance comparison of classification accuracy of missing modality on avMNIST dataset. $\eta\%$ represents the rates of complete modalities included.}
\resizebox{\linewidth}{!}{
\begin{threeparttable}                
\begin{tabular}{>{\arraybackslash}p{2.75cm}|>{\centering\arraybackslash}p{1cm}>{\centering\arraybackslash}p{1cm}>{\centering\arraybackslash}p{1cm}>{\centering\arraybackslash}p{1cm}}
\toprule
Method & 100\% & 70\% & 50\%  & 20\% \\ \midrule
Uni-Image & 87.3&- & - &- \\
Uni-Audio & 82.5 &-& - & - \\ \midrule
AutoEncoder$^\dagger$~\cite{baldi2012autoencoders}  &- &- & - & 88.8 \\
GAN$^\dagger$~\cite{goodfellow2020generative} & - &- & - & 89.5 \\
Full2miss$^\dagger$~\cite{shen2019brain} & -& - & - & 92.6 \\
SMIL$^\dagger$~\cite{ma2021smil}  &97.6& 96.4 & 95.6 & 94.4 \\
ShaSpec$^\dagger$~\cite{wang2023multi} &98.2 & 97.3 & 96.0 & 94.6\\
MCKD$^\dagger$~\cite{wang2024enhancing}&- & - & - & 95.1\\
Concatenate$^\ddagger$  &92.7  &88.1  &84.7  & 84.2 \\
MDA (Ours)$^\ddagger$  &\textbf{98.4} & \textbf{98.1} & \textbf{96.2} & \textbf{95.9} \\ \bottomrule
\end{tabular}
\begin{tablenotes}
\item[$^\dagger$] Only audio modality is missing. 
\item[$^\ddagger$] Missing modality at random.
\end{tablenotes}
\end{threeparttable}}
\label{table:mnist}
\end{table}

\begin{table}[!t]
\centering
\caption{Model performance comparison of classification accuracy of missing modality on MM-IMDb dataset. Evaluating performance using F1 sample and F1 Macro scores. $\eta\%$ represents the rates of complete modalities included.}
\resizebox{\linewidth}{!}{
\begin{threeparttable}
\begin{tabular}{>{\arraybackslash}p{2cm}|>{\centering\arraybackslash}p{0.7cm}>{\centering\arraybackslash}p{0.7cm}>{\centering\arraybackslash}p{0.7cm}|>{\centering\arraybackslash}p{1.2cm}>{\centering\arraybackslash}p{1.2cm}}
\toprule
& \multicolumn{3}{c|}{F1 sample} & \multicolumn{2}{c}{F1 Macro} \\ \cline{2-6}
& 100\% & 70\% & 20\% & 30\% & 30\% \\ \cline{2-6} 
Method &  &  &  & w/o Img & w/o Text  \\ \midrule
Uni-Image &36.1 & - & - & - & -  \\
Uni-Text &39.8 & - & - & - & - \\ \midrule
MFAS~\cite{perez2019mfas}&63.1 & - & - &-  & - \\
CentralNet~\cite{vielzeuf2018centralnet}&63.9  & - & -  &-  &- \\
ViLT~\cite{kim2021vilt} &64.6 & - & - &-  &- \\ \midrule
SMIL$^*$~\cite{ma2021smil} & - & - & 54.1  &- &- \\
MAPs$^\dagger$~\cite{lee2023multimodal} & - & - & - &46.3 &39.2 \\
MuAP$^\dagger$~\cite{dai2024muap} & - & - & - &47.2 &41.3 \\
MSPs$^\dagger$~\cite{jang2024towards} & - & - & - &47.4 &38.3 \\
Concatenate$^\ddagger$ &66.2 & 63.7 & 49.6 & \multicolumn{2}{c}{40.4}\\ 
MDA (Ours)$^\ddagger$ &\textbf{71.8} &\textbf{70.1} & \textbf{54.6} &\multicolumn{2}{c}{\textbf{47.7}} \\ \bottomrule
\end{tabular}
\begin{tablenotes}
\item[*] Only text modality is missing.  
\item[$\dagger$] Fixed missing modality. Different prompts are set for different missing modalities.
\item[$\ddagger$] Missing modality at random.
\end{tablenotes} 
\end{threeparttable}}
\label{table:movie}
\end{table}

The experimental results for both the avMNIST and MM-IMDb datasets, as detailed in Tables~\ref{table:mnist} and~\ref{table:movie}, underscore the robustness and adaptability of MDA. In the avMNIST dataset, MDA excels across all levels of data availability, from 20\% to 100\%, outperforming traditional methods like Concatenate and AutoEncoder, as well as state-of-the-art techniques such as SMIL and ShaSpec. This success is attributed to MDA’s ability to dynamically allocate weights to available modalities, ensuring high classification accuracy even with limited data. A similar trend is observed in the MM-IMDb dataset, where MDA achieves the highest F1 sample and F1 Macro scores, maintaining superior performance with missing modality, particularly when 70\% of modalities are randomly included. Notably, MDA also demonstrates a clear advantage over methods like MFAS, CentralNet, and ViLT which depend on complete modalities. In the MM-IMDb dataset, the performance gap between MDA and methods tailored for fixed missing modalities, such as MAPs, MuAP, and MSPs, under random missing conditions, further emphasizes MDA’s adaptability and generalizability. Overall, MDA’s consistent and superior performance across varying conditions of data completeness highlights its potential for real-world applications, particularly in medical diagnostics where data integrity is often variable.

\subsection{Interpretability analysis}
\label{sec:interpretability}
To address Challenge 4: The Comprehensive Interpretability of Multi-modal Fusion, we have conducted a thorough analysis of the interpretability of multi-modal fusion on MDA from both a macroscopic clinical diagnostic prior perspective and a microscopic model perspective, focusing on the variations in multi-modal attention.

\begin{figure}[!t]
    \centering
    \includegraphics[width=\linewidth]{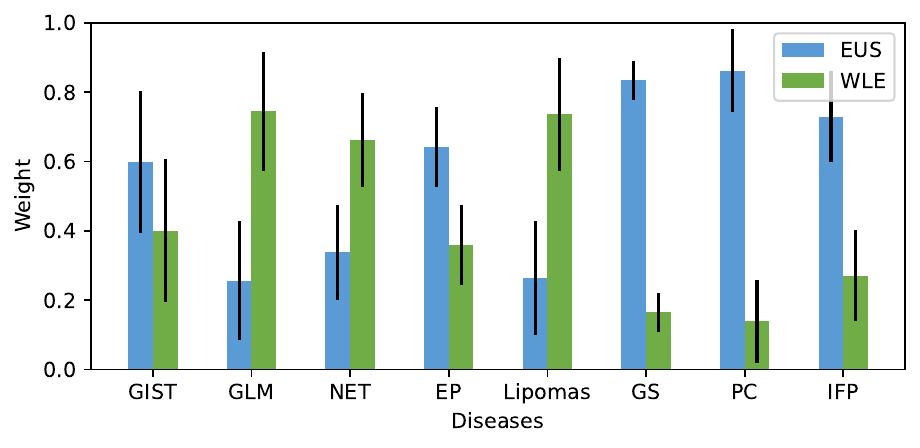}
    \caption{Macroscopic investigation of MDA weights for various diseases.}
    \label{fig:weight}
\end{figure}

\textbf{Interpretability in Diagnostic Standards Research}. 
Fig.~\ref{fig:weight} illustrates the average MDA weights of the proposed method for different disease categories. MDA weights $w_i$ for the modality $i$ can be calculated as in Eq.~\ref{eq:w_i}, 
where ${\mathbf{\Theta}_i}$ represents the MDA map, as discussed in Section~\ref{section2.4}.
\begin{equation}
w_i=\frac{\sum \mathbf{\Theta}_i}{\sum_{i=1}^{N} \sum\mathbf{\Theta}_i}
\label{eq:w_i}
\end{equation} 

MDA exhibits notable specificity in modal attention for different disease categories. It is important to note that no constraints were imposed during model training regarding which modalities should be attended to for the recognition of different diseases. We attribute this specificity to the proposed modality weight adaptation module working in conjunction with multi-disease classification. 
Furthermore, we analyzed the alignment between the model's specificity in modal attention for different disease categories and clinical priors. 
In clinical practice, multi-modal imaging techniques are frequently employed to gather comprehensive diagnostic information. For instance, the combination of EUS, WLE, CT, and MRI is utilized for the diagnosis of gastrointestinal diseases, with each imaging modality being sensitive to different features.
According to a study on gastrointestinal diseases~\cite{richardson1995well, chandler2019cochrane, jacobson2023acg}, using WLE alone to diagnose lipomas achieves a specificity of 99\% in clinical practice. Similar findings hold for NET. Conversely, GIST and EP rely more heavily on EUS ~\cite{jacobson2023acg}. This alignment between the MDA weights and clinical expertise is highly consistent. The findings can corroborate existing clinical experience and serve as a reference for modalities lacking such experience, which is precisely what is needed in the current field of multi-modal medical algorithms: reinforcing the role of artificial intelligence in establishing the gold standard for diagnostic use of multi-modality.

\begin{table}[!t]
\centering
\caption{Variations in MDA weights in missing modalities and learning with intrinsic noise.}
\resizebox{\linewidth}{!}{
\begin{tabular}{>{\centering\arraybackslash}p{1cm}|>{\centering\arraybackslash}p{1cm}|>{\centering\arraybackslash}p{1.5cm}>{\centering\arraybackslash}p{1.5cm}>{\centering\arraybackslash}p{1.5cm}>{\centering\arraybackslash}p{1.5cm}}
\toprule
Disease & Modal & Baseline & w/o EUS & w/o WLE & w/o Rep.\\ \midrule

\multirow{3}{*}{GIST} & EUS & 0.08$\pm$0.05 & \cellcolor{gray!25}0.11$\pm$0.06 & 0.06$\pm$0.06 & 0.27$\pm$0.12 \\
& WLE & 0.21$\pm$0.11 & 0.39$\pm$0.14 & \cellcolor{gray!25}0.05$\pm$0.06 & 0.40$\pm$0.15 \\
& Report & 0.72$\pm$0.13 & 0.49$\pm$0.13 & 0.90$\pm$0.10 & \cellcolor{gray!25}0.33$\pm$0.13 \\
\midrule
\multirow{3}{*}{GLM} & EUS & 0.17$\pm$0.10 & \cellcolor{gray!25}0.08$\pm$0.05 & 0.31$\pm$0.17 & 0.37$\pm$0.16 \\
& WLE & 0.26$\pm$0.12 & 0.40$\pm$0.13 & \cellcolor{gray!25}0.12$\pm$0.05 & 0.46$\pm$0.15 \\
& Report & 0.56$\pm$0.14 & 0.52$\pm$0.11 & 0.56$\pm$0.14 & \cellcolor{gray!25}0.17$\pm$0.09 \\
\midrule
\multirow{3}{*}{NET} & EUS & 0.33$\pm$0.10 & \cellcolor{gray!25}0.11$\pm$0.04 & 0.26$\pm$0.10 & 0.71$\pm$0.13 \\
& WLE & 0.10$\pm$0.07 & 0.35$\pm$0.11 & \cellcolor{gray!25}0.10$\pm$0.05 & 0.16$\pm$0.11 \\
& Report & 0.57$\pm$0.10 & 0.54$\pm$0.10 & 0.64$\pm$0.10 & \cellcolor{gray!25}0.13$\pm$0.07 \\
\midrule
\multirow{3}{*}{EP} & EUS & 0.16$\pm$0.08 & \cellcolor{gray!25}0.07$\pm$0.07 & 0.29$\pm$0.12 & 0.42$\pm$0.15 \\
& WLE & 0.38$\pm$0.17 & 0.34$\pm$0.17 & \cellcolor{gray!25}0.17$\pm$0.07 & 0.49$\pm$0.16 \\
& Report & 0.46$\pm$0.16 & 0.59$\pm$0.16 & 0.54$\pm$0.11 & \cellcolor{gray!25}0.09$\pm$0.06 \\
\midrule
\multirow{3}{*}{Lipoma} & EUS & 0.10$\pm$0.06 & \cellcolor{gray!25}0.02$\pm$0.01 & 0.02$\pm$0.02 & 0.61$\pm$0.14 \\
& WLE & 0.04$\pm$0.03 & 0.09$\pm$0.07 & \cellcolor{gray!25}0.01$\pm$0.00 & 0.15$\pm$0.11 \\
& Report & 0.86$\pm$0.07 & 0.89$\pm$0.07 & 0.97$\pm$0.02 & \cellcolor{gray!25}0.24$\pm$0.11 \\
\midrule
\multirow{3}{*}{GS} & EUS & 0.35$\pm$0.13 & \cellcolor{gray!25}0.13$\pm$0.08 & 0.33$\pm$0.12 & 0.77$\pm$0.09 \\
& WLE & 0.27$\pm$0.13 & 0.33$\pm$0.13 & \cellcolor{gray!25}0.27$\pm$0.06 & 0.12$\pm$0.08 \\
& Report & 0.39$\pm$0.13 & 0.54$\pm$0.13 & 0.40$\pm$0.14 & \cellcolor{gray!25}0.11$\pm$0.06 \\
\midrule
\multirow{3}{*}{PC} & EUS & 0.18$\pm$0.08 & \cellcolor{gray!25}0.07$\pm$0.03 & 0.12$\pm$0.06 & 0.62$\pm$0.12 \\
& WLE & 0.15$\pm$0.07 & 0.33$\pm$0.13 & \cellcolor{gray!25}0.06$\pm$0.05 & 0.25$\pm$0.09 \\
& Report & 0.67$\pm$0.12 & 0.59$\pm$0.12 & 0.82$\pm$0.05 & \cellcolor{gray!25}0.13$\pm$0.12 \\
\midrule
\multirow{3}{*}{IFP} & EUS & 0.14$\pm$0.08 & \cellcolor{gray!25}0.04$\pm$0.04 & 0.09$\pm$0.07 & 0.57$\pm$0.16 \\
& WLE & 0.13$\pm$0.10 & 0.19$\pm$0.10 & \cellcolor{gray!25}0.02$\pm$0.02 & 0.31$\pm$0.15 \\
& Report & 0.72$\pm$0.13 & 0.77$\pm$0.10 & 0.89$\pm$0.08 & \cellcolor{gray!25}0.12$\pm$0.06 \\

\bottomrule
\end{tabular}}
\label{table:3}
\end{table}

\textbf{Micro-visualization of model adaptation to missing modalities}
Table~\ref{table:3} demonstrates the variations in MDA weights for three modalities before and after incorporating missing modalities in the network.
The findings indicate that the proposed model is highly sensitive to these changes, as evidenced by the rapid and substantial shifts in modal-domain MDA weights across modalities when dealing with missing modalities. When EUS is missing, a significant decrease in EUS weights is first observed (0.34 to 0.05, 0.53 to 0.11), which is also observed when WLE is missing. When WLE is missing, MDA weights almost entirely shift towards the report modality, which is expected since the surface information of the tumor, reflected in WLE, cannot be found in EUS. When WLE is absent, the model cannot learn relevant knowledge solely from the EUS modality. However, the report modality contains descriptions of WLE and can serve as a substitute for the missing image modality. Hence, the weights assigned to the report modality increase. Similarly, in the absence of EUS, the emphasis shifts more toward the reporting modality rather than the WLE modality.  In the scenario where the report modality is missing, the redistribution of the reduced weight is approximately evenly allocated to both WLE and EUS to compensate for the absence. Due to the lack of information provided by the missing modality, experiments indicate that when a missing modality is present (i.e. when the modality does not provide information), the MDA module can effectively reduce the impact of the missing modality on the final results through weight redistribution.

\section{Conclusion}
The proposed MDA represents a pivotal advancement in multi-modal fusion, addressing the core challenges of missing modalities, intrinsic noise, and interpretability. MDA underscores the significance of a unified technical framework, which successfully constructs linear attention relationships between various modalities, permits adaptive control over modality weights, and scales to an arbitrary number of modalities with lower complexity.

This study pioneers the exploration of interpretability for key features and gold standard modalities in computer-aided diagnosis through the visualization of macroscopic multi-modal MDA maps. Its objective is to enhance the contribution of artificial intelligence to the standardization of diagnostic criteria. Furthermore, by visualizing the attention shifts in the internal correlations across multiple modalities at a microscopic level, the research imbues the model with self-explanatory capabilities.

{\small
\bibliographystyle{ieee_fullname}
\bibliography{MDA}
}

\end{document}